\documentclass[12pt,a4paper]{article}

\usepackage[utf8]{inputenc}
\usepackage[T1]{fontenc}
\usepackage[british]{babel}

\usepackage[margin=2.5cm]{geometry} 
\usepackage[onehalfspacing]{setspace}

\usepackage{amsmath}
\usepackage{amsthm}
\usepackage{amssymb}
\usepackage{tipa}
\usepackage{dsfont}
\usepackage{mathtools}
\usepackage{mathrsfs}
\usepackage{mathpazo} 
\usepackage{enumerate}
\usepackage{schemata}
\usepackage{latexsym}
\usepackage{amsfonts}
\usepackage[inline]{enumitem}
\usepackage{graphicx}
\usepackage{anysize}
\usepackage[font=footnotesize]{caption}
\usepackage{subcaption}
\usepackage{float} 
\usepackage{todo}
\usepackage[hyphens]{url}
\usepackage{hyperref}
\usepackage{verbatim}

\usepackage{tikz}
\usetikzlibrary{snakes}
\usetikzlibrary{decorations}
\usetikzlibrary{decorations.pathmorphing}
\usetikzlibrary{decorations.pathreplacing}
\usetikzlibrary{arrows,positioning}
\usepackage{listings}
\usepackage{changepage}

\newcommand*\xbar[1]{%
	\hbox{%
		\vbox{%
			\hrule height 0.5pt 
			\kern0.5ex
			\hbox{%
				\kern-0.1em
				\ensuremath{#1}%
				\kern-0.1em
			}%
		}%
	}%
}

\usepackage{titling}

\makeatletter 
\newcommand{\inspect}[1]{%
	\def\inspectspace{\mskip3mu\relax}
	\sbox\z@{$#1$}
	\sbox\tw@{\thickmuskip=0mu$#1$}
	\ifdim\wd\tw@<\wd\z@ \def\inspectspace{\mskip9mu\relax}\fi 
}
\makeatother 

\DeclareMathOperator{\Existsa}{\exists}
\newcommand{\Exists}[1]{
	\inspect{#1}
	\Existsa #1 \inspectspace
}

\DeclareMathOperator{\Foralla}{\forall}
\newcommand{\Forall}[1]{
	\inspect{#1}
	\Foralla #1 \inspectspace
}

\newcommand{\bool}{\texttt{bool}}

\newcommand{\Shift}{\texttt{Shift}}
\newcommand{\Day}{\texttt{Day}}
\newcommand{\Week}{\texttt{Week}}

\newcommand{\Event}{\texttt{Event}}
\newcommand{\EventList}{\texttt{EventList}}

\newcommand{\hours}{\text{h}}

\newcommand{\gshifts}{\texttt{shifts\_of}}
\newcommand{\gdays}{\texttt{days\_of}}
\newcommand{\gweeks}{\texttt{weeks\_of}}

\newcommand{\drivingTime}{\texttt{driving\_time}}

\newcommand{\totalTime}{\texttt{total\_time}}

\author{
	Bjørn Jespersen\\
	\small Universiteit Utrecht
	\and
	Ana de Almeida Borges\\
	\small Universitat de Barcelona
	\and
	Jorge del Castillo Tierz\\
	\small Universitat de Barcelona
	\and
	Juan José Conejero Rodríguez\\
	\small Universitat de Barcelona
	\and
	Eric Sancho Adamson\\
	\small Universitat de Barcelona
	\and
	Aleix Solé Sánchez\\
	\small Universitat de Barcelona
	\and
	Nika Pona\\
	\small Universitat de Barcelona
	\and
	Joost J. Joosten\\
	\small Universitat de Barcelona
}
\date{September 2018}
\title{When logic lays down the law}
\providecommand{\keywords}[1]{
	\small  
	\textbf{Keywords:} #1
}

\begin{document}
	
	\maketitle
	
	\begin{abstract}
    We analyse so-called computable laws, i.e., laws that can be enforced by automatic procedures. These laws should be logically perfect and unambiguous, but sometimes they are not. We use a regulation on road transport to illustrate this issue, and show what some fragments of this regulation would look like if rewritten in the image of logic. We further propose desiderata to be fulfilled by computable laws, and provide a critical platform from which to assess existing laws and a guideline for composing future ones.
	\end{abstract}	
	
	
	\newpage
	\tableofcontents
	\newpage

	\section{Introduction: the spirit, the letter and the logic of the law}
	\label{sec:introduction}

	\subsection{Law and language} 
	
	Legal texts are written in natural language, although of a regimented kind. The pragmatics of natural language is characterized by a principle of charity, which cuts speakers and authors a good deal of slack, because all parties to the discourse seek to cooperate toward getting the intended message across. Not so with legal texts, including, e.g., contracts, the signatories to which are likely to have countervailing interests; and tax laws, every loophole of which is liable to be used to lower the client's tax contribution. Therefore, the prose, apart from being crystal clear, ought to anticipate as many eventualities as possible, including far-fetched ones relying on degenerate or limiting cases, to offset the fact that the participants to the discourse are non-cooperative.

	One of the consequences of this is that legal texts seek to disambiguate certain sentential connectives. For a well-known example, the extensional connective `or' is ambiguous between inclusive and exclusive disjunction, so `and/or' is introduced to indicate that only inclusive disjunction is intended, as opposed to `either/or', which indicates an exclusive disjunction.
  
  Other connectives are no less ambiguous, but typically fail to receive disambiguation. An example would be the occasionally intensional connective `unless', as in `$A$, unless $B$', and cognates such as 'however'. If construed extensionally, `$A$, unless $B$' is truth-functionally equivalent to `$A$, if not $B$' ($\neg B \to A$), which is classically equivalent to  `$B$, if not $A$' ($\neg A \to B$), which would be the natural translation of `$B$, unless $A$'. If construed intensionally, i.e., as a connective between truth-conditions rather than truth-values, `$A$, unless $B$' will, for instance, fail to commute to `$B$, unless $A$' due to logical interdependence between the truth-conditions of $A$ and $B$.\footnote{Truth-conditions (truth-values-in-intension) are understood as per possible-world semantics, i.e. as satisfaction classes of possible worlds.}
  Commutativity fails, for instance, when $A$ and $B$ are indexed to two different times. The logical, as opposed to temporal, order between $A$ and $B$ may also matter as when, e.g., $A$ is a necessary condition for $B$ while $B$ is not a necessary (though perhaps a sufficient) condition for $A$.\footnote{
       As a simple example, let $A$ be the proposition that the sky is cloudy and $B$ the proposition that it is raining. Then $A$ is a (nomologically) necessary condition for $B$ and $B$ a (nomologically) sufficient condition for A.
  } \footnote{A logically transparent semantics for the different uses of `unless' and its cognates must specify in what ways `unless' connects $A$ and $B$. See \cite{Chandler1982} for a discussion of the temporal variant.}
	
  In the same vein, \cite{Robaldo2017} puts forward a notable example of logical and legal reasoning coming apart, of a similar type to the examples we will delineate in this article. The scenario is that whoever commits crime $X$ must spend two years in prison and whoever commits crime $Y$ must also spend two years in prison, and that a particular individual, $a$, has committed either $X$ or $Y$. Logic dictates that $a$ must spend two years in prison. But legal reasoning does not sustain this conclusion. The reason is that in the absence of evidence that $a$ committed $X$ or that $a$ committed $Y$, $a$ cannot be convicted of either crime and, therefore, cannot be sentenced to two years in prison.\footnote{In this particular case we see that \emph{constructive} logic actually follows reasoning about evidence much more closely than \emph{classical} logic that uses the \emph{tertium non datur}.}
	
	Moreover, legal texts seek to define their key terms or notions, albeit in prose. What motivates this move toward an exactitude not found in colloquial natural language is that legal texts strive to approximate a clarity often associated with mathematical, logical and scientific texts, which are similarly prone to abandon the principle of charity to avoid errors and biases.\footnote{
		In \cite{Haack2018}, Haack discusses some differences between scientific and legal endeavours, stressing that the latter are much more procedure-bound than the former are method-bound: `I would say, rather, that the purpose of a trial is to arrive at a determination of guilt, responsibility, punishment, liability, or whatever, in a legally-correct way' (ibid., 8), because evidence that is factually relevant may be procedurally inadmissible.
	}
	Without a high degree of exactitude, a legal text will be too imprecise and open to far more interpretations and inferences than was intended by the lawmakers.\footnote{
		We are proceeding on the assumption that legal contexts are not purposefully vague or designed with loopholes.
	}
	In this paper we shall provide examples from a particular EU Regulation pertaining to the driving and rest periods of truck and bus drivers as recorded by analogue or digital tachographs. It is particularly important to make sure there are no ambiguities or uncertainties in this case because the law is being enforced by automatic procedures, implemented in computers. These algorithms must know exactly how to proceed in every single situation, not only the most common ones. We found that this regulation is often unclear, which is unacceptable. It means that the people making decisions on what is legally correct are not the legislators, but the computer programmers. Hence we claim that legal texts of this sort - called computable -  should have the clarity and precision of a mathematical text. One could even require that the algorithm or its implementation be considered the official law. We discuss this possibility in Section \ref{sec:radical}.

	\subsection{Law and logic}
	
	When a mathematician, logician or programmer approaches a legal text, they might find themselves frustrated. The fact is that legal texts are literally designed by a committee, one composed of multiple politicians and lawyers and lobbyists, and must often absorb various compromises, amendments, derogations, qualifications, etc., thus turning the texts into patchworks. This fact does not in itself make legal texts inoperative for the purposes of day-to-day legal proceedings, for court decisions will privilege particular interpretations, but it does make them non-susceptible to logical or algorithmic treatment. That being the lay of the land, what are we to do when it is precisely our ambition to make legal texts susceptible to logical and algorithmic treatment?
	
	Here is the idea. Suppose we were to turn the table around between logicians and lawmakers. Suppose logicians would lay down the law by composing the actual legal texts. Then what might a logically rigorous legal text look like? This is the question we address in this paper. We are emphatically not claiming that our way must be the only way. We are claiming that a logically rigorous legal text will make it possible to reason from well-understood premises such that, potentially, the \textit{full informational} content of a given text can be extracted. Of course, such an approach makes sense only in certain well delimited domains that allow for disambiguation and for which such disambiguation is actually desirable. As such, we are currently mostly interested in laws with clear ontologies\footnote{
		In criminal law, intent is one of the general classes of \emph{mens rea} (mental elements) that plays a part in determining whether an act is a crime. For instance, in Canadian law intent is defined (by the ruling in \cite{RvMohan}) as ``the decision to bring about a prohibited consequence.'' However, it is questionable to what extent intent has a clear ontology. Intent in an action may be hard to determine, may or may not be subject to self-denial by the perpetrator, it may be pitted against contrary intentions, etc. A coin toss, although it may have ambiguous outcomes, can be unproblematically simplified to `heads' and `tails'. The same cannot be said of intent attributed to an act or an agent, which is always an issue to be determined, and may hinge on circumstantial qualities such that it cannot be broken down systematically into a clear ontology.
	} 
	that have a strong quantitative aspect. In particular we care about laws that are meant to be enforced by automatic procedures. It is only in these domains that substantial contributions of logic to the legal corpus are to be expected.
	
	A logical approach will make it possible to check the \textit{meta-theoretical} properties of the text, such as consistency, completeness, and feasibility (i.e., solvability and checkability with the use of a reasonable amount of computational resources like time and memory). In particular, it will be possible to issue a guarantee that a given legal corpus is logically consistent.\footnote{
		We are suppressing a philosophically, semantically and logically relevant discussion here. Proponents of true contradictions may want to allow, or even encourage, some legal corpora to include inconsistencies, though without `explosion' (i.e.~without \textit{ex contradictione quodlibet} being valid). One motive is the descriptive one that actual legal texts are bound to contain inconsistencies and the logical modelling should reflect that. Another is that inconsistencies are a fact of reality, for which there should be made room in the law. (See \cite{Priest1998}.) However, we are not going to presuppose that reality contains inconsistencies, nor do we want to include the category of true contradictions.
	}
	Such meta-theoretical properties are required of a logically tractable legal text. In Section \ref{sec:desiderata} we put forward a list of \textit{desiderata}. We should stress that in this paper we are not envisioning some universal logico-mathematical verdict-producing machine. Rather, what we have in mind is the ability to check a law for desirable formal properties before even implementing the law in order to arrive at verdicts. In the case of most laws, arriving at a verdict is a discretionary decision that requires not only deduction but also factual knowledge and reasoning by analogy and precedents.
	
	\subsection{Law and computation: a case study}
	
	Our \textit{modus operandi} is to develop a \textit{domain-specific formal ontology}, within which to lay down definitions of key notions from which to draw inferences. This way the bulk of the knowledge that can be extracted from the law will be inferential knowledge.\footnote{Inferential knowledge refers to knowledge that has been obtained as the conclusion of a sound deductive argument. Although this knowledge has been obtained via inference rather than, say, one's own observation or testimony, the knowledge may concern empirical facts.}
	
	Our material point of departure is a handful of paragraphs from Regulation 561/2006 enacted by the European Parliament on 15 March 2006 \cite{Regulation561}, henceforth `the Regulation'.\footnote{
		Our reason for bringing up this particular regulation is that the authors Bjørn Jespersen, Ana Borges, Eric Sancho, Aleix Solé, Nika Pona and Joost J. Joosten are all involved in a research and development project funded by Formal Vindications S.L.~and sponsored by the Generalitat de Catalunya. Formal Vindications S.L.~develops legal software for, \emph{inter alia}, the road transportation sector.
	}
	Though the spirit of this law is both stated fairly clearly in its Introduction and easily read off the paragraphs themselves, the letter of the law is still flawed. In this introduction we briefly sketch the most blatant of the problems, which share a common origin.
	
	The Regulation is basically about when, especially for how long, individual truck and bus drivers are obliged by law to rest, and when, especially for how long, they are allowed to drive on the road. As the Regulation is concerned with who does what and when along a time axis, the single most critical notion is the one of \emph{time}, as partitioned into atomic time units and assembled into aggregates of time units, including both interrupted and uninterrupted aggregates. Not to put too fine a point on it, the Regulation's understanding of time and its partitions and assemblies is hopeless. This flaw propagates through notions like \emph{period}, \emph{rest period} and \emph{driving period}.\footnote{
    For example, the notion of \emph{driving period} is ill-typed, see Section \ref{sec:driving_period}.
  }
	
	To be more specific, it is a major problem how \textit{week} is to be understood. The Regulation explicitly defines a week as the period between 00:00 on a Monday and 24:00 on the following Sunday.\footnote{
		Although `00:00' and `24:00' would seem to be mere notational variants co-designating the midnight hour (12 o'clock at night), they are not, as `00:00' designates the beginning of the day (a 24-hour cycle) and `24:00' the end of the day. That is, they are co-extensional but not co-intensional terms. In other words, even though `24:00' of one day $d_i$ and `00:00' of $d_{i+1}$ (the following day) designate the same moment in time, one expression is not interchangeable for the other in all contexts, as the former marks the end of a day and the latter the beginning of a day.
	}
	From the moment it is defined thus, `week' cannot also mean `any consecutive 168-hour period', which appears to be its usual meaning outside the Regulation. The problem is, the Regulation needs both the defined and the usual meaning of `week', but cannot have both on pain of homonymy. Furthermore, how are we to understand the adjective `weekly'? It usually means something like `occurs once a week'. In the light of the definition of \emph{week}, does `weekly' mean `occurs once between any Monday at 00:00 and the following Sunday at 24:00' or rather `occurs once within any 168-hour cycle'? These two interpretations are not equivalent. Consider an event starting at, say, 23:30 on a Sunday and finishing at 00:30 on the following day, which is a Monday and therefore the beginning of the week. If such an event consistently takes place within that span of time, it qualifies as being `weekly' in the second, but not the first, sense. Which week (in the defined sense of \textit{week}) does the event belong to?
	
	Still another problem is that not all intervals have been delimited on both sides. For instance, \S8.6 of the Regulation demands that a so-called \textit{reduced weekly rest period} must have been compensated by an extra rest period before the end of the third week after it. Hence, this paragraph would allow (in letter, if not in spirit) a compensation that had taken place, say, three years earlier. This loophole is symptomatic of a poor conception of time intervals. This example may be artificial, for sure, but imagine a situation where a driver knows that they have extra work lined up for the following week: are they allowed to anticipate by compensating in the form of additional rest in the week prior to it? This scenario would definitely tally with the spirit of the law. Unfortunately, humans do not have a `sleep savings account' to deposit excess sleep on, so this form of compensation seems to fall short of the objective of the Regulation.
	
	By turning the table around, designing a small legal fragment in the image of logic, we demonstrate one workable way in which a legal text may look like that lends itself to computation. The motivation for computability is that computable laws are known to be desirable in areas of professional transportation, such as automated and self-driving vehicles \cite{doi:10.2105/AJPH.2016.303628}. Our project is prescriptive rather than descriptive, as we are not offering interpretative strategies for existing legal texts, but are instead putting forward a way to apply a formal model for future legal texts to fit into and a critical platform from which to assess existing laws. 
	
	The rest of the paper is structured as follows. Section \ref{sec:naive_ontology} reconstructs the naïve ontology implicit in the Regulation. Section \ref{sec:flaws} charts some relevant shortcomings of the Regulation. Section \ref{sec:logic_to_law} sets out the conceptual foundations of our positive proposal; this includes putting on the table the problem of regular software as a non-reliable source of information (and its solution) in Section \ref{sec:software:reliability}. Section \ref{sec:desiderata_metatheory} offers a list of desiderata applying to logically and mathematically satisfactory legal texts, and it also examines the meta-theoretic properties of the framework put forward in Section \ref{sec:logic_to_law}. Finally, Section \ref{sec:implementation} presents examples of how to proceed when formally writing a law, specifically the specification of the law.

	\section{A reconstruction of the naïve ontology of Regulation 561}
	\label{sec:naive_ontology}
	
	We begin with a survey of the paragraphs of the Regulation which are relevant to our points in case, namely \S\S6-9. These paragraphs are basically intended to describe when drivers are allowed to drive and when they must rest. They follow a number of paragraphs that serve to define a host of notions, such as \textit{driver}, \textit{rest}, \textit{vehicle}, etc. Here we have reconstructed the naïve ontology presupposed and deployed in the Regulation.\footnote{
		We are using `naïve' in the philosophical sense of being based on unreflected common sense.
	}
	It is a straightforward ontology of \textit{who}, \textit{what} and \textit{when}, that is, the ontology concerns mainly driving and rest during time intervals, thus what is relevant is exclusively what the drivers do and when.
	
	\begin{description}
		\item[WHO:] \textit{Agents}. The only Agents are:
		\begin{itemize}
			\item \textit{Individual bus or truck drivers};
      \item \textit{Teams of bus or truck drivers} (usually two people).\footnote{
          This is referred to as \textit{multimanning}, but we will largely ignore this possibility, as it does not lead to any new insights. 
        }
		\end{itemize}
		Legal agents such as transportation companies are mentioned in the Introduction to the Regulation. Law enforcement (police officers) and courts (lawyers and judges) are presupposed rather than mentioned.
		
		\item[WHAT:] \textit{Events, activities}. The four fundamental ones are:
		\begin{itemize}
			\item \textit{work} (which is tantamount to \textit{driving});
			\item \textit{other work} (e.g. bookkeeping);\footnote{ Other work includes all activities as defined as working time in \S3a of Directive 2002/15/EC \cite{Directive15} except driving, including any work for the same or another employer, within or outside of the transport sector.}
      \item \textit{availability} (being available for driving);
			\item \textit{rest} (either for recuperation (\textit{break}) or for free disposal).
		\end{itemize}
		\item[WHEN:] \textit{Periods, intervals}. 
		\begin{itemize}
			\item \textit{sums} (i.e.~\textit{unordered periods} or \textit{intervals});
			\item \textit{sequences} (i.e.~\textit{ordered periods} or \textit{intervals}).
		\end{itemize}
		The Regulation does not stipulate a smallest time unit. However, Commission Implementing Regulation (EU) 2017/799 \cite{Regulation799} implementing Regulation (EU) No 165/2014 \cite{Regulation165} states that `(42) Time measured shall have a resolution better than or equal to 1 second', so for this reason we shall assume that seconds are the smallest time unit\footnote{
      Regulation 165/2014 \cite{Regulation165} and its implementation \cite{Regulation799} legislate the use of recording equipment in road transport.
  }. A given measure of time instantiates one of four properties:
		\begin{itemize}
			\item \textit{being work};
			\item \textit{being other work};
      \item \textit{being availability};
			\item \textit{being rest}.
		\end{itemize}
		To tie the elements of the various categories together, we shall also assume that each time unit is associated with an agent. Thus a given driver instantiates during a given moment exactly one of three states:
		\begin{itemize}
			\item \textit{working} (i.e.~\textit{driving});
			\item \textit{doing other work};
      \item \textit{being available};
			\item \textit{resting}.
		\end{itemize}
		The Regulation imposes various constraints on the permissible quantities (maximal values) of work and other work within three sorts of intervals:
		\begin{itemize}
			\item \textit{days};
			\item \textit{weeks};
			\item \textit{pairs of consecutive weeks}.
		\end{itemize}
	\end{description}

	The Regulation imposes no constraints (maximal values) on rest. But the drivers' employers all impose \textit{minimal values} on work (and perhaps also on other work). The Regulation is intended to balance these two countervailing interests such that drivers and their employers can all earn a living and the latter enjoy decent working conditions (i.e.~avoid being exploited) while not compromising road safety by being overworked.

	\section{Various fundamental flaws of the Regulation}
	\label{sec:flaws}

	Below is a catalogue of the fundamental flaws of the Regulation. Most of them are a consequence of flaws in keyword referents, as well as ambiguous uses of language.\footnote{This section is in part inspired by \cite{Marti2018} and \cite{Marti2016}.}
	
	\subsection{Technical language and references}
	One shortcoming concerns \textit{nouns, adjectives} and \textit{adverbs} formed from the same root. For instance, a \textit{driver} may not be \textit{driving}, i.e.~be behind the wheel, but be only a potential driver by being available to take over the wheel. There is also a potential source of confusion with `to work', which is ambiguous between \textit{work}, which is tantamount to \textit{driving}, and \textit{other work}, which is distinct from \textit{driving}; \textit{other work} would then be tantamount to \textit{other driving}, which is undesirable.
	Some clearly identifiable problems include:
	
	\begin{itemize}
		\item \textit{Period} is not defined, but it is used to characterize both \textit{duration} and \textit{interval} (see Section \ref{subsec:period}). \item \textit{Daily rest period} (\S4g) is poorly defined, as neither \textit{period}, nor \textit{daily} has been defined (so this amounts to hand-waving). \item In (\S4h), \textit{weekly rest period} does not tally with \textit{week} in (\S4i).
		\item	(\S4o) allows for one hour of solitary multimanning, which is a contradiction in terms. It is also left open what qualifies a non-driving driver as being `available for driving', e.g. whether other work is excluded by being so available, or whether an available driver may be resting.
		\end{itemize}
	
		\subsection{What is a period?}
	\label{subsec:period}
	
	Despite being a key notion, $period$ goes undefined. Furthermore, the several different uses of the term `period' are jointly incompatible. For example, on one occasions `period' needs to mean `uninterrupted interval' to make sense, which would suggest that an unqualified `period' could possibly be a union of intervals. Indeed, it is stated explicitly that a rest period can be an interrupted period (even though it remains unclear whether rest periods qualify as periods, as we have seen above). However, `period' also occurs in `each period of driving', which would be superfluous if the union of any two periods would also be a period (and it is reasonable enough to assume that a set of periods is closed under union). This leaves one wondering whether `is a period of driving' has (or is assumed to have) a defined meaning of its own. 
	
	\subsection{What is a driving period?}
  \label{sec:driving_period}
	\textit{Driving period}, in turn, \textit{is} defined, but it remains unclear whether `driving period' and `period of driving' are synonymous. Moreover, \textit{driving period} as defined is ill-typed. The definition identifies the driving period of a driver as the driver's accumulated driving time, where `driving time' means `the duration of driving activity recorded [by the tachograph]'. This would turn a driving period into a duration, thus contradicting the previous meaning of `period'. One could try to argue that \textit{driving period} is completely divorced from \textit{period}, despite sharing the word `period'. Still, the definition of \textit{driving period} states that a driving period `may be continuous or broken', which are two adjectives that do not apply to durations.

	\subsection{Implicit knowledge and ambiguity}
	Here we present a related linguistic issue, which will be further addressed in Section \ref{sec:desiderata} (Desiderata). We argue by means of the following example that substantial differences in meaning should not heavily depend on implicit knowledge of how the grammar of a language works. 
	
	The fundamental flaw that the following fragment exemplifies is that the literal meaning of the paragraph is nonsensical, for reasons that we are going to make explicit. To make sense of the paragraph the interpreter must assume that the verb phrase 'shall have taken' was not supposed to be written in future perfect, in the first place. 
	
	To begin, let us now consider \S8.2:
	
	\begin{quote}
		Within each period of 24 hours after the end of the previous daily rest period 
		or weekly rest period a driver shall have taken a new daily rest period.
		
		If the portion of the daily rest period which falls within that 24 hour period is at 
		least nine hours but less than 11 hours, then the daily rest period in question shall 
		be regarded as a reduced daily rest period.
	\end{quote}
	
	As a first stab, 'shall have taken' \textit{prima facie} amounts to the daily rest period having both begun and ended, since the tense is the future perfect. However, under this interpretation, once the daily rest period ends, a new period of 
	24 hours will have begun, within which another daily rest period `shall have taken' place, and so on.
	Therefore, a weekly rest period can never be undertaken (not even by extending a daily 	rest period to a weekly rest period, as specified in \S8.3), since in such a case the driver is unable to comply with \S8.2, given that it is not possible to fit in more than 24 hours\footnote{A weekly rest period as defined necessarily lasts more than 24 hours.} within a 24-hour period.
	But not undertaking weekly rest periods is a violation of \S8.6, which demands that the driver must take weekly rest periods. Moreover, again under this interpretation, the wording of 'which falls within that 24 hour period' essentially refers to a situation which could not be any different, and thus becomes irrelevant.
	
	To sum up, the literal interpretation of the fragment would (a) be inconsistent with \S8.6, and (b) render \S8.3 irrelevant. Therefore, the wording of 
	\begin{quote}
		If the portion of the daily rest period which falls within that 24 hour period is at 
		least nine hours but less than 11 hours, then the daily rest period in question shall 
		be regarded as a reduced daily rest period.
	\end{quote}
	leads us to believe that `shall have taken' must not be interpreted literally. 
	Interpreting `shall have taken' as stating that the daily rest period has at least begun immediately solves these problems and strikes us as being the likely intended meaning of this fragment of the Regulation.

  Interestingly, the Portuguese translation of the Regulation\footnote{
    European legislation is translated into all the official languages of the European Union.
    }
  does not have this problem, simply stating that the driver must take a new daily rest period inside of each 24 hour period after the end of the previous daily or weekly rest period.
  This illustrates an obvious question: how to make sure that all different translations of a legal text coincide? We propose a (perhaps too radical) solution in Section \ref{sec:radical}.

  	\subsection{Open-ended definitions}
	Exceptions are poorly expressed. The basic problem is how to handle open-ended inductive definitions (i.e.~definition by enumeration of kinds of cases without a final `nothing else' clause). This is encoded by locutions such as `$A$; however, $B$' and `$A$, unless $B$'. On a direct and uncharitable reading, the paragraphs that include exceptions are inconsistent, because they start with a universal claim that is followed by an exception, which amounts to this conjunction: $\forall x \phi(x) \wedge \exists x \neg \phi(x)$. Needless to say, something different is intended, but what? 
	
	Consider \S6.1:
  \begin{quote}
    The daily driving time shall not exceed nine hours.

    However, the daily driving time may be extended to at most
    10 hours not more than twice during the week.
  \end{quote}
  This article appears to have the overall structure of `$A$, unless $B$', where $A$ is expressed by `... shall not ...', which on a first approximation goes into: $\forall x \square \neg (\hdots x \hdots)$, while $B$ is expressed by `... however ...', which on a first approximation goes into: $\exists x \mathord{\Diamond} (\hdots x \hdots)$. Paradox ensues. One may wonder whether the $A$ clause expresses a universal proposition. It definitely does not talk about one particular or unique daily driving time. Yet it cannot be universal due to the $B$ clause, which would contradict the universal one. The problem arises when $A$ is read as a stand-alone condition. The way to look at it is that $A$ occurs embedded, just as it would in `If $A$, then $C$', so it cannot be factored out as a stand-alone clause on pain of paradox.
	
	\S6.1 is intended to express two simultaneous conditions on daily driving time. A correct formulation would be this: `The daily driving time must never exceed 10 hours, and the number of times that the daily driving time exceeds 9 hours must be at most two per week.' Here, an extensional (i.e.~not progressive) conjunction has replaced the problematic `however' (similarly for `unless', presumably).

	\subsection{Placement of weekly rest periods}
	
	Let us consider a limiting case implied by the following clauses of \S8 of the Regulation.
	
	\begin{itemize}
		
		\item[\S8.1.] A driver shall take daily and weekly rest periods.

		\item[\S8.9.] A weekly rest period that falls in two weeks may be
		counted in either week, but not in both.
		
	\end{itemize}
	
	A first step is to understand the terms involved inasmuch as they have been established earlier in the text:
	
	\begin{itemize}
		
		\item[\S4(h)] `regular weekly rest period' means any period of rest of at least 45 hours.
		
  	\item[\S4(i)] `a week' means the period of time between 00.00 on a
		Monday and 24.00 on the following Sunday.
	\end{itemize}
	
\begin{figure}[H]
		\centering
		\begin{tikzpicture}
		\draw (1,0) -- (13,0) ;
		
		\draw (1,-0.25) -- (1,0.25) ;
		\draw (3,-0.25) -- (3,0.25) ;
		\draw (5,-0.25) -- (5,0.25) ;
		\draw (7,-0.25) -- (7,0.25) ;
		\draw (9,-0.25) -- (9,0.25) ;
		\draw (11,-0.25) -- (11,0.25) ;
		\draw (13,-0.25) -- (13,0.25) ;

		\node[align=center, below] at (2.55,-0.1) {45h};
		\node[align=center, below] at (5.1,-0.1) {48h};
		\node[align=center, below] at (7.4,-0.1) {72h};
		\node[align=center, below] at (9.3,-0.1) {45h};
		\node[align=center, below] at (11.5,-0.1) {45h};
		
		\node[align=center, above] at (1,0.2) {A};
		\node[align=center, above] at (3,0.2) {B};
		\node[align=center, above] at (5,0.2) {C};
		\node[align=center, above] at (7,0.2) {D};
		\node[align=center, above] at (9,0.2) {E};
		\node[align=center, above] at (11,0.2) {F};
		\node[align=center, above] at (13,0.2) {G};
		
		\draw[snake, segment amplitude=1.5, segment length=1.5] (2.4,0) -- (2.7,0);
		\draw[snake, segment amplitude=1.5, segment length=1.5] (4.8,0) -- (5.4,0);
		\draw[snake, segment amplitude=1.5, segment length=1.5] (6.8,0) -- (8,0);
		\draw[snake, segment amplitude=1.5, segment length=1.5] (9.1,0) -- (9.5,0);
		\draw[snake, segment amplitude=1.5, segment length=1.5] (11.3,0) -- (11.7,0);
		\end{tikzpicture}
    \caption{Six consecutive \textit{weeks} and five weekly rest periods (serpentine lines) taken by a hypothetical driver.}
  \label{fig:WRP1}
	\end{figure}
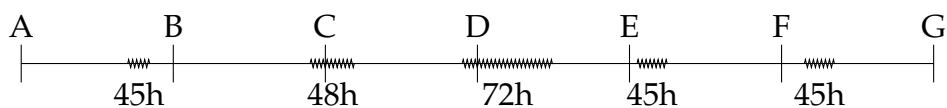

	Now, let us consider the situation depicted in Figure \ref{fig:WRP1}. Each segment denotes a week, with time flowing from left to right. Furthermore, each serpentine line denotes time spent in a weekly rest period.
	
	Since two of these weekly rest periods fall between two weeks, it is reasonable to want to find a procedure that will determine whether there exists a way of counting each of them within one week or the other, such that the situation becomes legal. That there should exist a procedure to determine the legality of the situation is crucial, and such a procedure is listed among our desiderata section (Section \ref{sec:desiderata}). It may be possible in more complex situations for there to be legal combinations, but no terminating computable procedure to determine them. Consequently, depending on how the law is laid down, epistemic `holes' are likely to crop up.
	
	In our simplified case, a seemingly reasonable algorithm would be to check if each week segment has a weekly rest period. Segment $AB$, for example, unambiguously has one of 45 hours. $BC$ depends on where we count the on-the-fence 48h hours; as we want each week to have a weekly rest period, we count these hours as falling within segment $BC$. In $CD$, however, we encounter the same problem: hence, the 72-hour weekly rest period is counted as falling within $CD$. 
	However, now segment $DE$ does not have any weekly rest period, so a fine is issued. The problem is that this line of decision-making, which at first glance seems reasonable, issues a fine to the driver for not having taken a weekly rest period on week $DE$, that is, for not having rested enough during that specific week. This is in blatant contradiction with our intuitions. The driver has been fined for not resting \textit{enough} during the very week of the controlled period in which they have rested the most!
	
We may alternatively try an algorithm that would attempt to shift weekly rest periods to the left by default, and, in case of a contradiction, try shifting to the right, and, in case of further contradictions, shift the previous unshifted weekly rest periods, and so on. The problem here is, whichever procedure one chooses, how to guarantee that a situation does not become so contrived that it flies in the face of the spirit of the law? How to avoid situations that contradict our intuitions about what the legislators intended to consider as illegal?
	
	Here, the Regulation does not pose a logical problem, nor is it inconsistently worded. But logic is not entirely unrelated to this issue. The complexity that results from \S8.9 generates a potential combinatorics problem, and our \textit{prima facie} intuition of the consequences may not withstand analytic scrutiny. It is in the analysis that logic comes in. In Section \ref{sec:desiderata} we outline specific goals in law-drafting that can contribute to stave off complexity issues.

\subsection{Non-locality via long weekly rest periods}

  In this and the next subsection we illustrate a very undesirable property of the law, which we name \emph{non-locality}. 
	
  Assume a driver starts their first weekly rest period at midday on a Sunday and ends it on the next Saturday. By \S8.9, this rest corresponds to the first week\footnote{
    Recall that as per \S4(i), ` ``a week'' means the period of time between 00.00 on Monday and 24.00 on Sunday.'
  } (the one in which the driver started the weekly rest), and so the driver would be fined because they did not rest enough during the second week (although, in fact, they rested for almost a week).

Indeed, they did something wrong: not resting enough on the \emph{first} week, but the reason the police officer provides when giving the fine is `You have not rested during the \emph{second} week'.

However, if the driver takes a new weekly rest period starting on the next day (Sunday) and ending it on the following Saturday as before, the second week turns to 'legal' whereas the third week is the illegal one. In other words, because the driver did not rest enough during the first week, the law states that the third week should be fined.

We could keep repeating the same reasoning and extending the schedule one week at a time, but that would result in a large amount of needless pages. In general, the driver can get fined on Week $n$ because of the lack of rest that actually happened on Week $1$. Moreover, each of the intermediate weeks is legal, since there is another weekly rest that can be assigned to them.

The problem arises when we have a large $n$. We know that tachograph records must be saved for a year, but they may be erased after that. Thus, let us assume a driver performs the considered schedule (having the weekly rests from Sunday to Saturday) for a year and then the following week does not start a new weekly rest. We know that the situation would be illegal, but the existing records would not back this, implying that the schedule was legal. Since there could be any number of weeks between the wrongdoing and the moment it is detected, we conclude that the Regulation is not local.

\subsection{Non-Locality via compensations}
	\label{subsec:locality}
	
  In this section we give\footnote{In this paper we are rather sketchy about our proofs. Similar results are proven in \cite{Castillo2018} and have actually been formalized in Coq.} a different example of non-locality. 
  In particular we show that there are two weeks $A$ and $B$ that are each by themselves legal but together are not legal. But that is not all. We show that we can find such $A$ and $B$ so that there are $n$ many weeks in between them where $n$ can be an arbitrarily large natural number. Moreover, we see that Week $A$ followed by all the $n$ weeks \emph{is} legal just as the $n$ weeks followed by $B$ are legal. The question then arises: \emph{where is the illegality?} It is in the combination between $A$ and $B$, where $A$ and $B$ can be arbitrarily far apart from each other. Clearly this is not a good feature for a law.

  Our point of departure is \S8.6:
	\begin{quotation}
		\noindent
		In any two consecutive weeks, a driver shall take at least:
		\begin{itemize}
			\item Two regular weekly rest periods, or
			\item one regular weekly rest period and one reduced weekly rest period of at least 24 hours.
		\end{itemize}
		However, the reduction shall be compensated by an equivalent period of rest taken en bloc before the end of the third week following the week in question.
	\end{quotation}

We give an intuition of why the law is non-local. The main idea is well illustrated in the following example. Generally we will only consider situations concerning weekly rest periods in which the driver has rested at least 24 hours.
	
\begin{figure}[H]
		\centering
		\begin{tikzpicture}
		\draw[line width=0.5mm](0,0)--(2,0);\draw[line width=0.5mm](2,0)--(10,0);
		\draw[line width=0.5mm](10,0)--(12,0);
		\foreach \x/\xtext in {0,2,10,12}
		\draw(\x,10pt)--(\x,-10pt);
		\foreach \x/\xtext in {4,6,8}
		\draw(\x,10pt)--(\x,-10pt);
		
		\node[text width=1cm] at (1.25, 0.35) {A};
		\node[text width=1cm] at (1.25,-0.25) {44};
		\node[text width=1cm] at (3.25,-0.25) {45};
		\node[text width=1cm] at (5.25,-0.25) {45};
		\node[text width=1cm] at (7.25,-0.25) {45};
		\node[text width=1cm] at (9.25,-0.25) {24};
		\node[text width=1cm] at (11.25,-0.25) {45};
		\node[text width=1cm] at (11.25, 0.35) {B};
		\end{tikzpicture}
    
    \caption{Illegal interval of six consecutive weeks done by a hypothetical driver.}
    \label{fig:example}
	\end{figure}
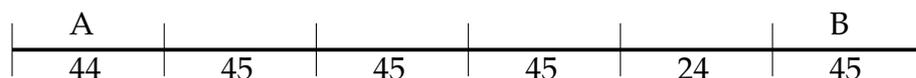

  Throughout this subsection, line segments represent weeks, and the numbers attached to them represent the number of hours rested during each week. In Figure \ref{fig:example}, the first and last segments represent the weeks $A$ and $B$ we mentioned before.
	
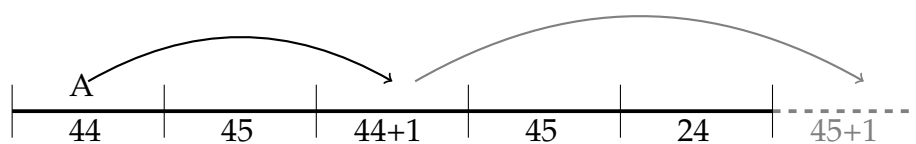
\begin{figure}[H]
		\centering
		\begin{tikzpicture}
		\draw[line width=0.5mm](0,0)--(2,0);\draw[line width=0.5mm](2,0)--(10,0); \draw[gray, line width=0.5mm, dashed](10,0)--(12,0);
		\foreach \x/\xtext in {0,2}
		\draw (\x,10pt)--(\x,-10pt);
		\foreach \x/\xtext in {4,6,8,10}
		\draw(\x,10pt)--(\x,-10pt);
		\foreach \x/\xtext in {12}
		\draw[gray](\x,10pt)--(\x,-10pt);
		
		\draw[->,thick, yshift=2ex]  (1,0) to [out=30,in=150] (5,0);
		\draw[->,thick, gray, yshift=2ex]  (5.3,0) to [out=30,in=150] (11.2,0);

    \node[text width=1cm] at (1.25, 0.35) {A};
		\node[text width=1cm] at (1.25,-0.25) {44};
		\node[text width=1cm] at (3.25,-0.25) {45};
		\node[text width=1cm] at (5,-0.25) {44+1};
		\node[text width=1cm] at (7.25,-0.25) {45};
		\node[text width=1cm] at (9.25,-0.25) {24};
		\node[gray,text width=1cm] at (11,-0.25) {45+1};
		\end{tikzpicture}

    \caption{First five weeks of the example represented in Figure \ref{fig:example}, together with a possible sixth week that would make the whole interval legal.}
    \label{fig:most_example}
	\end{figure}

  As shown in Figure \ref{fig:most_example}, if we do not consider the last week, the remaining interval is rendered legal by the law, for we can assume that the hours to be compensated will be incorporated in the week we took out.
	
\begin{figure}[H]
		\centering
		\begin{tikzpicture}
    \draw[line width=0.5mm](2,0)--(10,0);
    \draw[line width=0.5mm](10,0)--(12,0);
    \draw[gray, line width=0.5mm, dashed](12,0)--(14,0);

		\foreach \x/\xtext in {2,4,6,8}
		  \draw(\x,10pt)--(\x,-10pt);
		\foreach \x/\xtext in {10,12}
		  \draw (\x,10pt)--(\x,-10pt);
    \draw[gray](14,10pt)--(14,-10pt);
		
		\draw[->,thick, gray, yshift=2ex]  (8.9,0) to [out=30,in=150] (13,0);

		\node[text width=1cm] at (3.25,-0.25) {45};
		\node[text width=1cm] at (5.25,-0.25) {45};
		\node[text width=1cm] at (7.25,-0.25) {45};
		\node[text width=1cm] at (9.25,-0.25) {24};
    \node[text width=1cm] at (11.25,0.35) {B};
		\node[text width=1cm] at (11.25,-0.25) {45};
    \node[gray, text width=1.5cm] at (13.2,-0.25) {45+21};

		\end{tikzpicture}

    \caption{Last five weeks of the interval represented in Figure \ref{fig:example}, together with a possible sixth week that would make the whole interval legal.}
    \label{fig:rest_example}
	\end{figure}
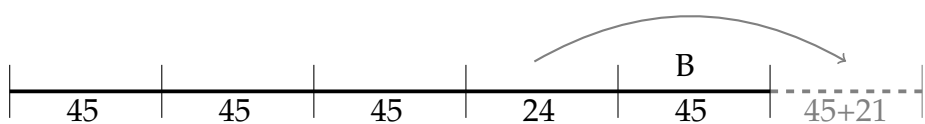

  Similarly, if remove week $A$ from the example in Figure \ref{fig:example}, the resulting interval (represented in Figure \ref{fig:rest_example}) is also legal, since we can assume that the compensation for the fourth week takes place in the weeks outside our interval.

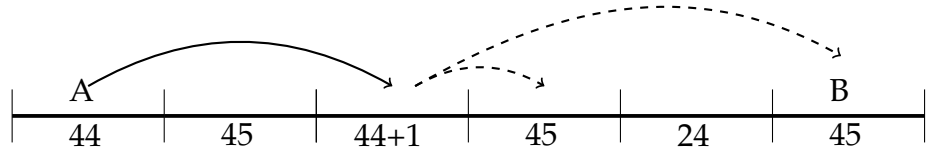
\begin{figure}[H]
		\centering
		\begin{tikzpicture}
		\draw[line width=0.5mm](0,0)--(2,0);\draw[line width=0.5mm](2,0)--(10,0);
		\draw[line width=0.5mm](10,0)--(12,0);
		\foreach \x/\xtext in {0,2,4,10,12}
		\draw(\x,10pt)--(\x,-10pt);
		\foreach \x/\xtext in {4,6,8}
		\draw(\x,10pt)--(\x,-10pt);
		
		\draw[->,thick, yshift=2ex]  (1,0) to [out=30,in=150] (5,0);
		\draw[->,thick, yshift=2ex, dashed]  (5.3,0) to [out=30,in=150] (7,0);
		\draw[->,thick, yshift=2ex, dashed]  (5.3,0) to [out=30,in=150] (10.9,0.4);

    \node[text width=1cm] at (1.25,0.35) {A};
		\node[text width=1cm] at (1.25,-0.25) {44};
		\node[text width=1cm] at (3.25,-0.25) {45};
		\node[text width=1cm] at (5,-0.25) {44+1};
		\node[text width=1cm] at (7.25,-0.25) {45};
		\node[text width=1cm] at (9.25,-0.25) {24};
    \node[text width=1cm] at (11.25,0.35) {B};
		\node[text width=1cm] at (11.25,-0.25) {45};
		\end{tikzpicture}

    \caption{The same interval of Figure \ref{fig:example}, with an attempt to assign compensations (dashed lines) that ultimately fails.}
    \label{fig:example_illegal}
	\end{figure}

	However, the interval of Figure \ref{fig:example} is illegal, as Figure \ref{fig:example_illegal} illustrates. This is because after compensating the first week according to article \S8.6, we still have to compensate one hour, but we cannot allocate it within any of the three following weeks without having two consecutive reduced weekly rest periods.
	
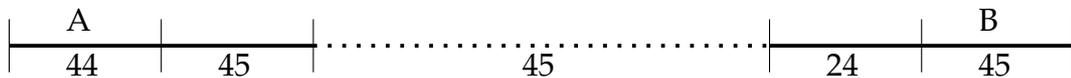
\begin{figure}[H]
\centering
\begin{tikzpicture}
\draw[line width=0.5mm](0,0)--(2,0);\draw[line width=0.5mm](2,0)--(4,0); \draw[loosely dotted, line width=0.5mm] (4,0) -- (10,0);\draw[line width=0.5mm](10,0)--(12,0);\draw[line width=0.5mm] (12,0) -- (14,0);
\foreach \x/\xtext in {0,2,12,14}
    \draw(\x,10pt)--(\x,-10pt);
\foreach \x/\xtext in {4,10}
    \draw(\x,10pt)--(\x,-10pt);
\node[text width=1cm] at (1.25,-0.25) {44};
\node[text width=1cm] at (1.25,0.35) {A};
\node[text width=1cm] at (3.25,-0.25) {45};
\node[text width=1cm] at (7.25,-0.25) {45};
\node[text width=1cm] at (11.25,-0.25) {24};
\node[text width=1cm] at (13.25,-0.25) {45};
\node[text width=1cm] at (13.25,0.35) {B};
\end{tikzpicture}

\caption{General example of an illegal interval that is legal when week A or week B is erased.}
\label{fig:general_example}
\end{figure}

  More generally, for any sufficiently large $n$ we can find an interval of weeks which is non-local. The correspondent interval (illustrated in Figure \ref{fig:general_example}) has a similar structure to the one we have treated. The first week has a 44 hour weekly rest period, and all the following weeks have 45 hour weekly rest periods except for the penultimate one, which has a 24 hour weekly rest period.

In this situation if we take out one of the weeks A or B the remaining interval will be legal, but the interval as it stands is illegal. 
	
	\section{From logic to law}
	\label{sec:logic_to_law}

	\subsection{Reverting the point of view}
	The remedy we propose is, first of all, to formally conceptualize the Regulation as a knowledge base embedded within a specific domain. Lawyers and judges are conceptualized as agents who query the law for answers to questions that fall within the domain assigned to the law \cite{Duzi2018}. A question that falls outside the domain is considered ill-conceived and as such will be dismissed by the law. Conceptualizing the law as a knowledge base serves to operationalize the law: it contains explicit knowledge, which can be read off immediately, and when rules of inference are applied to explicit knowledge the implicit knowledge of the law can be teased out and converted to explicit knowledge possessed by the inquiring agent. The full informational value of a law is understood to be the union of its explicit and implicit knowledge.
	
	By modelling a law as a knowledge base, we make legal texts amenable to the same techniques that are also applied to multi-agent systems, within which cooperative agents exchange information in order to increase their base of explicit knowledge. Individual pieces of information may be either factual or analytic (the latter being definitional or semantic). The purpose of the law is to distribute knowledge among the human agents; it is a source of knowledge that agents can tap into. The basic format of the interaction is that agents ask the law questions and the law answers them, provided they fall within its compass as circumscribed by its ontology.
	
	The questions that can be put to the law form a small taxonomy, in terms of \textit{wh-} questions, yes/no questions, and alternative questions, which dictates the logical type the answer to a given question must have. For instance, the answer to the question, `Is a driver allowed to drive for ten consecutive hours if entering a time zone lying further east?' must be either `Yes' or `No' (formally, 1 or 0, or $\top$ or $\bot$, i.e.~a truth-value), and cannot be, say, `42' (which is instead the right sort of answer to a request for the extension of a magnitude) or, `Multi-manning can, under certain circumstances, be done by a solitary driver' (which is an answer to a totally different question, whose answer must be either a truth-value or a proposition). Analytic questions aside (which serve to clarify concepts), the underlying question that is being put to the Regulation will always be whether X is legal, where X is a particular action consuming a particular amount of time. Questions can be posed in any of the three manners described above.
	
	The law should be a particular kind of knowledge base, namely one that is in flux. It cannot be static but must be dynamic for the simple reason that laws are revised. Some portions are revoked, other portions are added (with no guarantee that the resulting corpus will be consistent). When a law is a knowledge base in flux, agents within the system are required to keep abreast of such revisions. This may involve discarding obsolete knowledge, as something that used to be legal/illegal has now become illegal/legal. One may still know that a time-indexed proposition was true (thanks to `eternalisation'), but one cannot know that the same proposition, only without time-indexing, is now true, for the revisions of the law have rendered the proposition false. Eternalisation is the way to preserve the monotonicity of legal knowledge (`knowledge is never lost'), but as laws continue to be revised one's knowledge becomes historical. Without eternalisation, it is obvious that legal knowledge must be non-monotonic.
	
	Our sophisticated (as opposed to naïve) ontology recycles the \textit{who} and the \textit{what} of the naïve ontology above, but it replaces the \textit{when} with a more sophisticated account of time and temporal intervals.\footnote{
		We adopt Unix time, such that the zero hour is 00:00:00 UTC, 1 January 1970.
	}
	\textit{Time} can be thought of in at least three different ways:
	\begin{enumerate*}[label={(\roman*)}]
		\item as an \textit{instant} (a null or positive measure);
		\item as an \textit{interval}, either precisely or loosely circumscribed (e.g. from dusk to dawn), which may again be either discontinuous or uninterrupted; if uninterrupted, it will be modelled by a concave function on an interval; if discontinuous, if will be modelled by a finite union of concave intervals; 
		\item as the \textit{duration}, i.e.~cumulative length, of an interval, which will be modelled by a magnitude like \textit{the number of planets}, here applied to magnitudes typed to take sets of seconds to a number (their cardinality).
	\end{enumerate*}
	
	\subsection{Software reliability}
	\label{sec:software:reliability}
	When logic lays down the law, the result should be that the law becomes computable. In fact, laws specifically written to be enforced by automatic procedures already exist, and recently have a new-found stance within the realm of computer software \cite{Casanovas2008}. This includes laws designed to be computed without there necessarily being any human intervention whatsoever. For example, we have speeding laws, enforced automatically by radars, DNA sequencing programs,\footnote{
    An American federal judge has recently ordered that the code of a DNA sequencing program be made open-source, so that it could be checked for mistakes. This code was being used in court as evidence that certain people had been in certain places, and its correctness was put into doubt. See \cite{DNA} for more details.
  }
   etc. Let us suppose, for instance, that we have a law that has been drafted according to the parameters presented in this paper; such a law will, accordingly, be computable. Of course, the interpretation of a legal text will not be in the hands of the software industry. Software developers will be working from formal specifications (in logical and mathematical form), and all they will have to do is design a program that follows these specific instructions. This covers the engineering or technical aspect of the road code.
	
	However, we cannot expect all software to be bug-free \cite{McConnell2004}. This means that some of the software running a computable law will eventually output wrong evidence that could be the basis for an unfair verdict. This means that we are dealing with one additional layer of potential arbitrariness in law enforcement. It also means that the problem of legal responsibility for arbitrariness rears its head. Who is left with the responsibility of arbitrary law enforcement?
	
	It is unacceptable that software used to decide whether to fine someone should provide wrong data. Given this, we either reject software as a reliable source of judiciary information, or we find a way of writing infallible software. Fortunately, there is such a way: formally verified software. With the use of proof assistants such as Coq or Isabelle, it is possible to formally prove the correctness of computer code, and to automatically extract software. This software is the result of logico-mathematical proofs developed by humans and checked by computers\footnote{
    One may complain that if the proofs are written by humans and checked by computers, then they are in principle no more trustworthy than code written directly by humans. Even though it is true that a possibility for error remains, it is much, much smaller. Proof assistants are especially made with this concern in mind. See \cite{trustCoq} for a breakdown of what can go wrong with Coq.
  }
    \cite{SoerensenUrzyczyn2006}. In other words, the resulting code is the transfiguration of an exact mathematical proof. Therefore, it becomes a reliable source of correctness, insofar as mathematics is the most exact science we have, and consequently amounts to a reasonable, if tedious, solution.
	
	\subsection{A radical solution}
  \label{sec:radical}

  We stated that laws should be written in a clear and unambiguous way, satisfying also some other requirements (see Section \ref{sec:desiderata}). However, such a task would not solve all of the problems described in Section \ref{sec:flaws}.

  Computable laws are laws amenable to being enforced by automatic procedures. But there is always a loss in translation from a natural language text to an algorithm. Our proposal that the law is written in a mathematical language seeks to ameliorate this problem. However, there is a more drastic solution: have the algorithm be the law.
  In that case, it is not that there would be no natural language text explaining the algorithm, but simply that the onus would be on the algorithm. It would be a reversal of roles.

  We will not take this idea further in this paper but let us briefly mention some considerations on such a radical viewpoint. Technically speaking this algorithmic turn would solve many problems on interpretations, thereby taking out ambiguities and possibly saving costs spent on courses divulging the spirit and interpretation of the law. On the other hand, by doing so, it will make the law more rigid. Since algorithms are entirely deterministic and unambiguous, any plea against an application of the law can only consist of a plea against the law itself, thereby disabling the whole useful machinery of jurisdiction. This is a loss of flexibility and versatility.

  There is also a more fundamental objection. The law should in principle be accessible to everyone. That is to say, any literate person should be able to access, read, and take note of a law's purpose and intention. If the law were an algorithm, then only programmers or engineers would be able to do so, thereby making legal consultation only available to certain elites. Indeed, far more people can read text than code. On the other hand, who can fully understand technical legal texts? Knowing how to read does not imply being able to grasp the purport of a technical legal text written in legal jargon.

  One can also argue that the law simply \emph{is not} or \emph{should not} be an algorithm. When writing the algorithm, this feels like an interpretation of a more fundamental law that is likely to be enunciated in natural language. However, a similar objection applies to a law formulated in natural language where one may maintain that this is just a linguistic projection of a collection of fundamental ethical intuitions and beliefs. The radical viewpoint would thus imply that law schools should include programming as an integral part of the curriculum.

The objection of code being unreadable certainly holds for the current specification languages of proof-checkers. However, much improvement on this is to be expected over the years to come where the languages should become more conceptual and versatile. One can even conceive of special purpose fragments of formal languages where certain ontologies (intervals, seconds, lists, etc.) are already pre-defined.

	\subsection{Examples}
	
	In 1990, the Dutch Tax and Customs Administration started a project whose purpose was to manage the entire process from creating legal texts to implementing them. The name of the project is POWER, for `Programme for an Ontology-based Working Environment for design and implementation of Rules and regulations'. Its aim was to extract knowledge from different juridical sources and implement it in order to reduce costs, time and effort.\footnote{More about POWER and its verification and validation techniques (called VALENS) can be found in \cite{Spreeuwenberg}.}
	
	Research was also done on automated norm extraction, i.e., extracting norms from law texts written in natural language \cite{Engers}.
	
	In Finance it is becoming more and more common to employ formal verification methods to ensure the safety, legal compliance and correctness of large volume trading processes. The growing complexity of financial algorithms and regulations makes it no longer possible to rely solely on an informal idea of why an algorithm fulfils its purpose. The state-space these algorithms operate on is so vast that a human agent could not possibly survey every scenario.
	
	For a current state of affairs we refer to \cite{Formal_verification_financial_algorithms}, where a general overview of how this formalization can take place is presented, and a more detailed description of what needs to be formalized in this field and why. As an example, the paper shows how a verification tool was developed by Aesthetic Integrations, examining the formal verification of a trading venue's matching logic, and trading system connectivity.
	
	Other interesting approaches include describing financial contracts as algorithms \cite{Composing_contracts}, in which an algebraic approach is pursued in order to create the specifications of financial contracts, leading to a much more reliable evaluation of the contracts. Even if this approach involves no formal verification, the gains in terms of rigour increase notably.
	Following this path, \cite{Certified_symbolic_management} creates a formal variant of what was initiated in \cite{Composing_contracts}. They present a symbolic framework, whose contract language is implemented in Coq, for modelling financial contracts, and they show how this framework is able to express derivatives such as foreign exchange derivatives, among others.

	\section{Desiderata and meta-theory}
	\label{sec:desiderata_metatheory}

  In this section we propose a list of desiderata to be followed by all (computable) legal texts, and give an overview of a model for formal properties of a law.
	
	\subsection{Desiderata}
	\label{sec:desiderata}
	As this is a position paper outlining how to generate legal texts, we have compiled a set of desiderata.
	
	\begin{description}
		\item[Unambiguous prose:]
		  If, contrary to Section \ref{sec:radical}, it is decided to write the law in natural language, two or more linguistically possible interpretations of the meaning of a given fragment must be avoided. Whenever it is desirable that multiple interpretations be considered equally valid they should be explicitly marked as different interpretations and explicitly conjoined with an inclusive disjunction: multiple interpretability should never hinge on an ambiguity in the wording.
		
		\item[Clearly circumscribed formal ontology:]
		A formal ontology delimits the sphere of what a legal text can `talk about' (and derivatively what it cannot `talk about'). A clearly circumscribed formal ontology comes with exact categories or types of entities and exact descriptions of how they are capable of interacting (and derivatively how they are not capable of interacting). Such an ontology makes it feasible to describe some cases as paradigmatic and others as limiting cases.
		
		\item[Decidability and feasibility:]
		We understand that a legal text must provide a procedure which, for any correct data describing an action within the domain contemplated by the text, can be used to decide whether the action is legal or not. In particular, it must have the property of locality described in Section \ref{subsec:locality}. This is what we call `decidability'. On the other hand, we call such a procedure `feasible' when the set of steps and conditions are computable using a reasonable amount of computational resources, such as time and memory.
		
		\item[Consistency and satisfiability:]
		We take it to be a non-negotiable requirement that a law must be logically consistent for the simple reason that an inconsistent law fails to specify what is legal and what is not. Furthermore, an inconsistency classically entails anything. It must also be satisfiable; this is, it cannot be the case that there are no possible actions an agent can take in order to follow the law.
		
	\end{description}
	
	\subsection{A model for formal properties} 

  The `mini-universe' circumscribed above is one that we construe as a knowledge base from which to obtain inferential knowledge. It is only fair that we should adumbrate the sort of logical machinery we consider suitable for agents that turn to the law for knowledge. So to be more specific, Section 5.1.5 of \cite{Duzi2010} provides the details of a logic of inferable knowledge. Given a stock of explicit knowledge and a logical intelligence (i.e., a set of valid rules of inference), an agent is in a position to extract new explicit knowledge from existing explicit knowledge by drawing inferences. The new knowledge is inferable, because the logic tracks what the agent could get to know were they to perform the relevant acts of inference. This logic comes with a fixed point at which no further inferences can be drawn. The two key mappings of this logic are one that takes a set of input propositions (namely, the existing stock of explicit knowledge) to a set of output propositions (namely, the inferable knowledge), which we refer to as the explicit-to-inferable knowledge function; and another from a set of premises (propositions) to a conclusion (a proposition), which serves to model the logical intelligence of a given agent. The formal properties of this epistemic logic include:
	\begin{description}
		\item[Sub-classicality:] If $\alpha$ is derived from a stock of knowledge $\Gamma$ then $\alpha$ is entailed by $\Gamma$. That is, if $f$ is the function taking $\Gamma$ to the set of its logical consequences then the result of applying the above function from explicit to inferable knowledge to $\Gamma$ is a subset of the result of applying $f$ to $\Gamma$.
		\item[Reflexivity:] $\Gamma$ is a subset of the result of applying the explicit-to-inferable knowledge function to $\Gamma$. 
	\end{description}
	
	These two features preserve the factivity of knowledge and are in keeping with the realistic assumption of real-world agents being resource-bounded (formally, by the inferable stock of knowledge being a subset of the set of entailments).
	
	Thanks to the above features, the logic also includes monotonicity:
	\begin{description}
		\item[Monotonicity:] If $\Gamma$ is a subset of $\Gamma'$ then the result of applying the explicit-to-inferable knowledge function to $\Gamma$ is a subset of the result of applying said function to $\Gamma'$.
	\end{description}
	
	The explicit-to-inferable knowledge function is assumed to not be idempotent, as it is used to compute only one inferential step at a time:
	\begin{description}
		\item[Lack of idempotence:] For some $\Gamma'$, the result of applying the function to the result of applying it to $\Gamma'$ is not a subset of the result of applying it to $\Gamma'$.
	\end{description}
	
	In a realistic set-up, different agents will have different (if perhaps overlapping) logical intelligences, which makes cooperation among agents desirable with a view to obtaining \emph{distributed} knowledge. Conversely, lack of cooperation among agents (perhaps due to conflicting interests) can be modelled as failure to obtain particular pieces of distributed knowledge. A quick example to fix ideas: assume that $a$ knows that the implication from $A$ to $B$ is true, whereas $b$ knows that $B$ is false. If they compare notes, they will be in a position to infer that $A$ must also be false. If they do not, they miss out on getting to know $\neg A$ on the basis of their respective explicit knowledge and logical intelligences.

	\section{Writing a specification}
	\label{sec:implementation}

	As was pointed out earlier, a legal text is more likely than not to have been designed by a committee of politicians, lawyers and lobbyists, with various compromises, amendments, derogations, qualifications, etc. Now imagine that this law is only enforceable by software, as is the case with many cyber-laws. It is one thing for lawyers or judges to disagree and argue about the `spirit of the law' in the case of these `patchwork' texts. It is quite another thing for a software engineer to be commissioned with the implementation of such a text. Since programming requires clear and fixed technical specifications, we must strike a compromise between natural language and computability (see Section \ref{sec:radical}). We deem logic to be the right sort of bridge between the two. For this purpose, to illustrate our approach, we provide an example specification of three articles of the Regulation, thus showing how a fragment of it could be written if logic had laid down the law. We also show how to check that it complies with the desiderata and meta-theoretical properties outlined in Section \ref{sec:desiderata_metatheory}.
	The basic structure of the process consists of the following steps: 
	
	\begin{enumerate}
		\item Thoroughly understand what is required of the regulation.
		\item Choose the ontology. \label{item:ontology}
		\item Formalize each statement of the regulation using logic and the ontology as building blocks.
		\item If the ontology turns out not to be ideal, go back to Step \ref{item:ontology} and continue from there.
		\item Having logically formalized every statement, check that the desiderata described in Section \ref{sec:desiderata} are heeded.
	\end{enumerate}
	
	\subsection{Main data types}
	\label{subsec:datatypes}
	
	To model time, we use an abstraction where the smallest unit is the second, and each moment is represented as a natural number. For almost all purposes in the Regulation the moment in time when something occurs (e.g.~May 4th 2017 at 10:37:28) is not what we are interested in; rather, we are interested in \textit{durations}.
	
	As the main building block in our ontology, we chose the notion of $\Event$, which is an interval of time (represented, for example, as a start time and a duration), together with information about what happened during that time (for example, the vehicle was moving and there were two people in the vehicle). There cannot be conflicting information in an $\Event$. For example, it is not possible that in the same $\Event$ the vehicle was both moving and standing still. In such a case we would need two separate $\Event$s to describe the situation.
	
	All the information recorded by a tachograph can be described as a list of consecutive $\Event$s. If such a list has some specific correctness properties, we call it an $\EventList$.

	We also use structures to represent $\EventList$s with some additional properties. Since we are interested in how much time goes by without the driver taking a rest of at least $45$ minutes, we use the notion of $\Shift$ to represent $\EventList$s that start and end with consecutive rests of at least $45$ minutes. Our notions for $\Day$s and $\Week$s work in the same manner. The $\gshifts$ an $\EventList$ are then the list of $\Shift$s that occur in that $\EventList$, and similarly for $\gdays$ and $\gweeks$.
	
	Finally, regarding the examples below, we should also mention the functions $\drivingTime$ and $\totalTime$, which calculate the time spent with the vehicle moving and in total in a given $\EventList$, respectively.
	
	\subsection{Restrictions}
	
	As an example, we present our high-level formalization of three restrictions from the Regulation. We took care that the formalization was simple enough to be implemented in Coq, and are in the process of doing so.
	
	\begin{enumerate}
		
		\item The daily driving time shall not exceed ten hours. 
		\begin{equation*}
		f_1 :=
		\lambda \ el : \EventList \ .
		\Forall{d \in \gdays(el)}
		\drivingTime(d) \leq 10 \hours
		\end{equation*}
		
		\item After a driving period of four and a half hours a driver shall take an uninterrupted break of not less than 45 minutes, unless he takes a rest period.
		\begin{equation*}
		f_2 :=
		\lambda \ el : \EventList \ .
		\Forall{s \in \gshifts(el)}
		\drivingTime(s) \leq 4.5 \hours
		\end{equation*}
		
		\item A weekly rest period shall start no later than at the end of six 24-hour periods from the end of the previous weekly rest period.
		\begin{equation*}
		f_3 :=
		\lambda \ el : \EventList \ .
		\Forall{w \in \gweeks(el)}
		\totalTime(w) \leq (6 \times 24)\hours
		\end{equation*}
		
	\end{enumerate}
	
	Thus, given an $\EventList$ corresponding to the recordings of a specific tachograph during a specific period of time, we require that it satisfies each of these restrictions (and many others). To see whether it does, we implement the functions above and apply them to our specific $\EventList$.

  Following the line of thought mentioned in Section \ref{sec:radical}, it would be possible to have (an extension of) this formalization as \emph{the} regulation. This would mean that the interpretation would be fixed, and it would not make sense to question whether the formalization was faithful to the Regulation. However, one could (and should) still ask whether the choices made in the formalization were faithful to its intended meaning.

	\subsection{Meta-properties}
	
	With this formalization at hand, it is simple to check whether the desired meta-properties hold. The prose is unambiguous, as we use a formal language to describe the restrictions. The ontology qualifies as being clearly circumscribed for the same reason.
	
	\begin{description}
		
		\item[Decidability and feasibility:]
		The decidability of each restriction follows from the decidability of $\bool$, which is already proven in Coq. As for complexity, we just have to see that each function is at most polynomial (in fact, linear) in the size of the input $\EventList$. Specifically, observe that $\drivingTime$ and $\totalTime$ can be defined to traverse the $\EventList$ once and perform only additions and subtractions. Then the main functions traverse the input once and perform boolean comparisons on the result of the $\drivingTime$s and $\totalTime$s.
		
		\item[Consistency and satisfiability:]
		In order to check that our restrictions are internally consistent and satisfiable, we have to show that there is an $\EventList$ that satisfies them. That is, we need to prove the following statement:
		\begin{equation*}
		\Exists{el : \EventList} (
		f_1 \ el
		\land
		f_2 \ el
		\land
		f_3 \ el
		).
		\end{equation*}
		The empty $\EventList$ suffices, though of course there are many examples of non-trivial $\EventList$s that abide by all of the restrictions.
		
	\end{description}
	
	\section*{Conclusion}
  \addcontentsline{toc}{section}{Conclusion}
	
	Above we asked what the desiderata are for a logically perfect corpus that lends itself to being computed. We put forward a set of meta-theoretic desiderata. We also showed how a fragment of an existing law will look like when recast in the image of logic, and proposed the more radical approach where a formalization is the law. This serves to provide a critical platform from which to assess existing laws, as well as a guideline for drafting future laws.

  \addcontentsline{toc}{section}{References}
	\bibliographystyle{acm}
	\bibliography{bib,articles}

	\todos
	
\end{document}